\documentclass{article}

\usepackage[preprint]{corl_2026} 
\usepackage{amsmath}
\usepackage{amssymb}
\usepackage{booktabs}
\usepackage{graphicx}
\usepackage{wrapfig}

\title{Unleashing Infinite Motion: Scaling Expressive Quadrupedal Motion via Generative Video Priors}

\newcommand{\samethanks}[1][\value{footnote}]{\footnotemark[#1]}
\author{
Youzhi Liu\\
Amap, Alibaba Group\\
\texttt{\fontsize{7.5}{9}\selectfont liuyouzhi.lyz@alibaba-inc.com} \\
\And
Li Gao\thanks{Corresponding authors.}\\
Amap, Alibaba Group\\
\texttt{\fontsize{7.5}{9}\selectfont liangliang.gl@alibaba-inc.com} \\
\And
Yifei Qian\samethanks\\
Amap, Alibaba Group\\
\texttt{\fontsize{7.5}{9}\selectfont qianyifei.qyf@alibaba-inc.com} \\
\AND
Liu Liu\\
Amap, Alibaba Group\\
\And
Yang Cai\\
Amap, Alibaba Group\\
\And
Wenbin Tang\\
Amap, Alibaba Group\\
}

\begin{document}
\maketitle

\begin{abstract}
 Quadruped robots have achieved remarkable locomotion, yet their behavioral repertoire remains confined to a few gaits—far from the expressive, companion-like presence long envisioned for them. Attempts to import the humanoid recipe of large-scale motion data have inherited one tacit assumption: that robot motion must first pass through an animal body, making data collection dependent on cooperative animals, reconstruction fragile across species, and retargeting ill-posed across incompatible morphologies. We propose Uni-Mo, a fully automated pipeline that removes the animal from the loop by reframing data scarcity as a generation problem: an LLM proposes motion prompts, a video diffusion model synthesizes the corresponding robot behaviors, and the generated videos are lifted into 3D reference trajectories used to train tracking policies deployed on a real Unitree Go2. To make naively-drifting generations reliably extractable, we introduce an Identity Consistency Loss that enforces appearance coherence across frames.
  We release Quad-Imaginarium at \url{https://github.com/Amap-Robotics/Quad-Imaginarium}, the resulting open-source dataset of
  7{,}488 language-annotated quadruped motions (18.5 hours) spanning
  acrobatic and performative behaviors. We validate 392 randomly sampled motions on a real Unitree Go2 with a 96.7\% deployment success rate, complemented by a 97.6\% success rate across the full dataset in simulation.                                    
\end{abstract}


\section{Introduction}
\label{sec:introduction}

Quadruped robots have come a long way: they run across rubble, climb stairs, recover from kicks, and traverse terrains unthinkable a decade ago~\citep{10610200, inproceedings, rudin2021learning, haarnoja2024soccer}. And yet, watch a state-of-the-art quadruped for long enough and a strange impression sets in: it moves beautifully, but it has almost nothing to say. Its entire behavioral repertoire collapses into a handful of gaits—walk, trot, run, jump~\citep{li2023terrainadaptive, yang2023vim}—repeated across every context and every task. This is not the robot the public imagines. The cultural archetype of a quadruped companion, embodied by characters like Doraemon, is not a faster locomotion machine but an expressive, animal-like presence capable of rich, situated interaction. Humanoid robots, fueled by tens of thousands of human motion clips, are already moving in this direction, learning to dance, gesture, and emote~\citep{cheng2024expressivewholebodycontrolhumanoid, fu2024humanplushumanoidshadowingimitation, 10.1145/3528223.3530110, ji2024exbody2, he2024hover, wococo2024, zhuang2024humanoidparkour}. Quadrupeds are not. This asymmetry raises a question we believe is overdue: \emph{can quadruped robots break out of locomotion and acquire a motion repertoire as rich and expressive as their humanoid counterparts?}

The answer hinges on data. Progress on humanoid robots has been driven by two complementary pipelines: large-scale optical motion capture~\citep{AMASS_CMU, AMASS, chen2022mld}, and in-the-wild video reconstruction built on parametric body models such as SMPL~\citep{smpl}, exemplified by systems like VideoMimic~\citep{allshire2025videomimic} and SLoMo~\citep{zhang2023slomo}. When the community has tried to bring the same recipe to quadrupeds, it has almost always inherited one tacit assumption: that robot motion must first be expressed in an animal body—captured from a real dog, reconstructed from a video of a cheetah, or fit to a parametric quadruped model, before it can be transferred to the robot. The animal, in other words, is treated as a topologically distant intermediate.

\begin{figure}[t!]
    \centering
    \includegraphics[width=\textwidth]{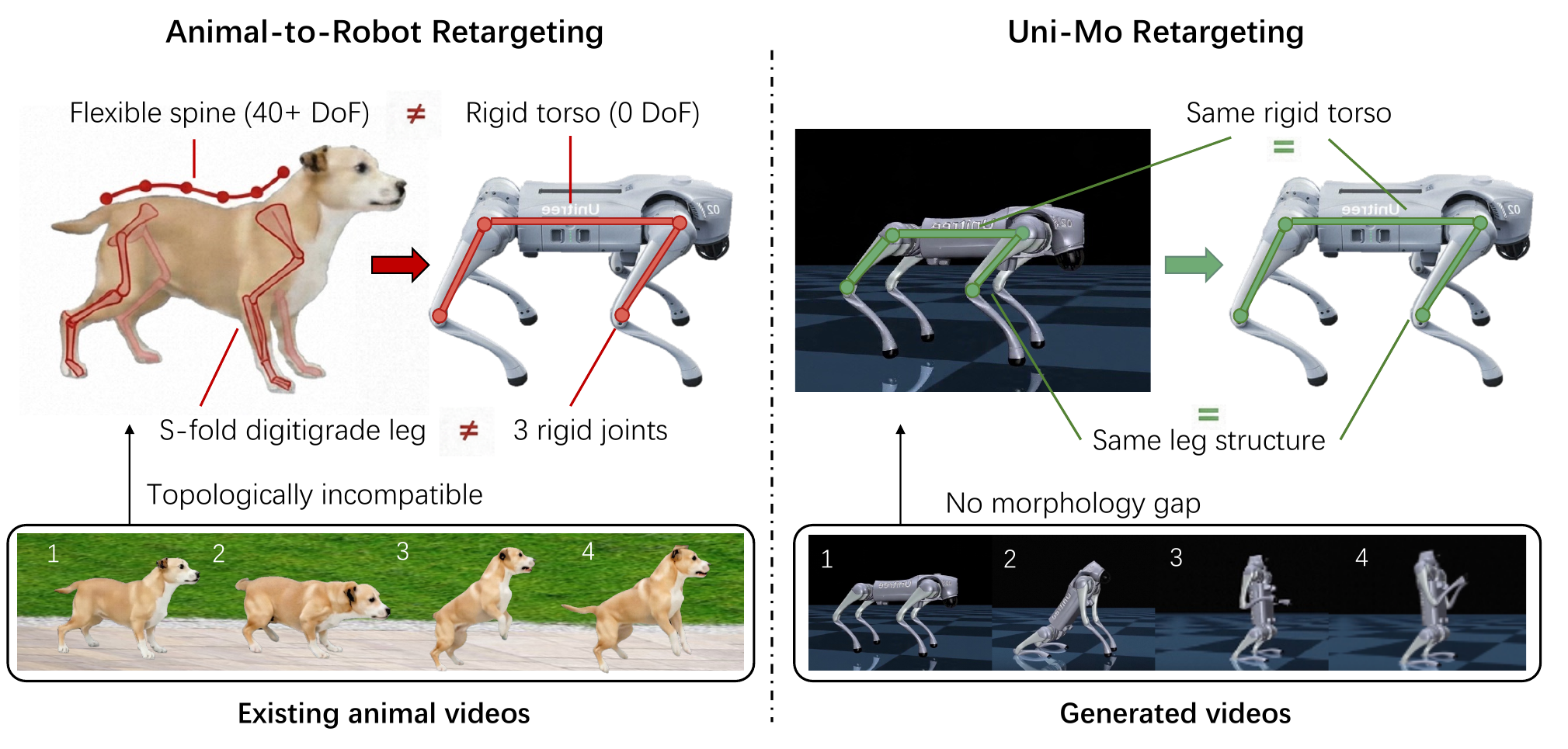}
    \caption{Animal-to-robot retargeting is ill-posed due to morphological mismatch (left); Uni-Mo sidesteps this by generating and extracting motion directly on the target robot (right).}
    \vspace{-20pt}
    \label{fig:teaser}
\end{figure}

This assumption looks natural, but it is precisely what makes the pipeline brittle. On the capture side, it ties data collection to the cooperation of real animals, who cannot be instructed to hold calibration poses or perform prescribed actions on cue; as a result, existing animal MoCap datasets~\citep{AcinoSet, wang2025dogmolargescalemultiviewrgbd, yu2021ap10k} remain narrow in both species and behavior, dominated by basic locomotion and missing precisely the expressive, companionship-oriented motions we care about~\citep{zhao2025aggressive, chanesane2024rlwav}. On the video side, it forces every clip through a species-specific 3D reconstruction step that relies on SMAL~\citep{8100069}, a parametric animal body model far less mature than its human counterpart SMPL~\citep{smpl}, and must contend with deformable surfaces, heavy self-occlusion, and large appearance variation across species and breeds. And even when reconstruction succeeds, the recovered motion lives on a skeleton that is topologically alien to the robot. Retargeting across this gap routinely yields physically infeasible motion, no matter how accurate the upstream reconstruction, as shown in Figure~\ref{fig:teaser}.

These are not three independent failure modes. They are three faces of the same unexamined commitment: that the path to robot motion must pass through an animal body. Once this commitment is named, an alternative becomes visible. If the animal intermediate is what makes data scarce, reconstruction fragile, and retargeting ill-posed, then the cleanest fix is not to engineer better animal models or better cross-morphology retargeters, it is to remove the animal from the loop altogether.

This is the premise of our work. To realize robot-centric generation, we turn to large-scale video diffusion models~\citep{wan2025, kong2024hunyuanvideo, seedance2026seedance20advancingvideo} as a synthetic source of motion. Pretrained on internet-scale video, they already depict a wide range of 
quadruped behavior when prompted~\citep{mou2025dimodiverse3dmotion}. The remaining challenge is to steer this 
generic prior toward a specific robot embodiment with enough visual fidelity 
that the generated videos can be reliably lifted into 3D reference trajectories through a pipeline of off-the-shelf perception components, at which point training a tracking policy becomes a standard problem.

We address both with Uni-Mo, a fully automated pipeline that fine-tunes a video diffusion model into a robot-aware generator, lifts its outputs into 3D reference trajectories, and trains a tracking policy deployable on a real Unitree Go2. The central technical obstacle along this path is what we call identity drift: over longer horizons, current video models deform the robot's body non-rigidly, shift its appearance, and distort its proportions across frames—violating the rigid-body assumption that any kinematic extraction depends on. To overcome it, we introduce an Identity Consistency Loss: a differentiable objective that enforces, frame by frame, that the generated robot remains the same rigid body with the same identity, and is backpropagated into the generator without eroding its underlying motion prior. To conclude, our main contributions are threefold:

1. We propose Uni-Mo, a fully automated pipeline that turns natural-language prompts into deployable quadruped policies, reframing data scarcity as a generation problem that scales with compute. We instantiate and validate the pipeline on a Unitree Go2.

2. We propose an Identity Consistency Loss, a generator-agnostic fine-tuning 
  objective that constrains every generated frame to remain consistent with the initial reference of the robot in both structure and appearance.

3. We release Quad-Imaginarium, a large-scale open-source dataset of 7{,}488 high-quality 3D reference motions totaling 18.5 hours, together with human-corrected action labels and language instructions, providing the community with a ready-to-use foundation for research on expressive quadruped motions.

\section{Related Work}
\label{sec:related_work}
 
\paragraph{Video Generation Models for Robot Motion Learning}
Video generation models have been explored for robot learning along three axes. For \emph{grounding into executable actions}, PhysWorld~\citep{mao2025physworld} reconstructs an interactive physical world from generated videos, while Gen2Real~\citep{ye2025gen2real} and GenMimic~\citep{ni2025genmimic} convert generated videos into deployable policies via physics-aware optimization or RL tracking. Nearly all target manipulation or humanoids, and physical plausibility is enforced through posterior reconstruction or policy tracking. Work directly targeting quadrupeds is scarcer: QuaDreamer~\citep{wu2025quadreamer} applies video generation to quadrupeds but for panoramic perception, while cross-morphology transfer methods such as TrajSkill~\citep{tang2025trajskill}, NovaFlow~\citep{li2025novaflow}, and NIL~\citep{albaba2025nil} still rely on explicit retargeting or keypoints and have not leveraged general-purpose video generation to discover motions at scale. Uni-Mo complements these by addressing an upstream problem---identity drift in the generated video---through an Identity Consistency Loss that turns a general-purpose video generator into a scalable quadruped motion engine without species-specific reconstruction.

\paragraph{Diverse Motion Learning for Quadruped Robots}
Quadruped motion diversity has historically been built on animal motion priors. \citet{peng2020imitateanimals} showed that imitating animal MoCap enables multiple gaits and dynamic jumps; AMP~\citep{peng2021amp} and its robotic extension~\citep{escontrela2022amp_robots} converted motion clips into style rewards, while VIM~\citep{yang2023vim} and Imitate and Repurpose~\citep{bohez2022imitate} organized priors into reusable motion libraries. To reduce MoCap dependence, SLoMo~\citep{zhang2023slomo} and STMR~\citep{yoon2024stmr} extract motions from monocular video, RLWAV~\citep{chanesane2024rlwav} uses action-classification scores from wild video as RL rewards, and~\citet{zhao2025aggressive} recover 3D motions from monocular animal footage.
On the synthetic-data front, DiffuseLoco~\citep{huang2024diffuseloco} learns multi-skill policies from offline data, and~\citet{chen2025inbetween} generate in-between gait sequences.
Recent dataset efforts QuadFM~\citep{gao2026quadfmfoundationaltextdrivenquadruped} and T2QRM~\citep{10.1145/3696409.3700230} augment locomotion data with teleoperation and artist authoring but still rely on real animal or human intermediaries. All of these remain confined to \emph{extracting} actions from existing recordings, bounded by what an animal happens to have been observed doing. Uni-Mo fills this gap by treating motion acquisition as a video \emph{generation} problem, scaling with compute rather than with the cooperation of real animals.

\section{Methodology}
\begin{figure}[t]
    \centering
    \includegraphics[width=\textwidth]{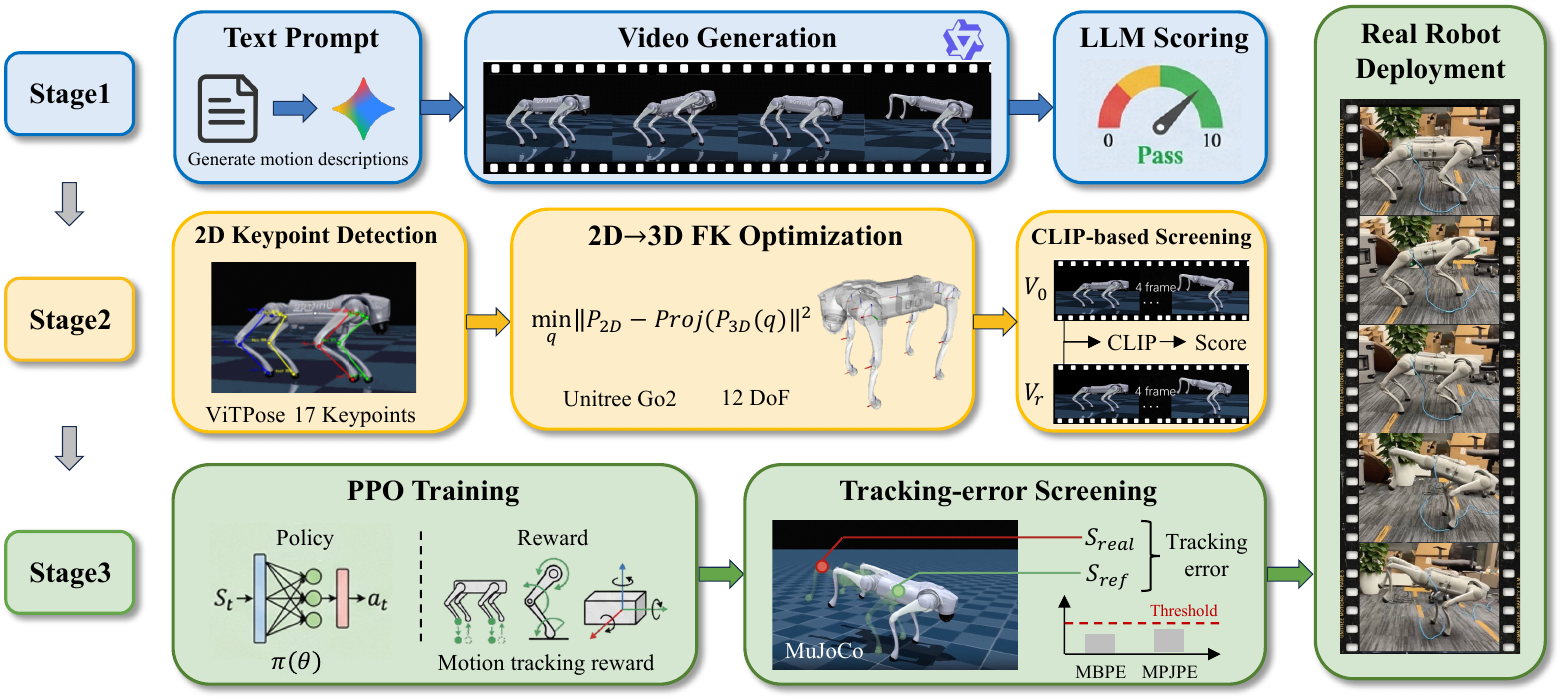}
    \caption{The Uni-Mo pipeline: natural-language prompts drive an identity-consistent video diffusion model, whose outputs are lifted into 3D reference trajectories and used to train PPO tracking policies deployed on a real Unitree Go2.}
    \label{fig:pipeline}
\end{figure}
\subsection{Pipeline Overview}
\label{sec:overview}

Uni-Mo turns natural-language motion prompts into deployable quadruped motions through a fully automated pipeline. We instantiate the pipeline on a Unitree Go2; the design itself makes no Go2-specific assumptions and applies to any quadruped with a known URDF. The pipeline has three stages.  (i) A large language model proposes a diverse set of candidate motion descriptions which, together with a fixed canonical reference image of the Go2, are passed to a video diffusion model fine-tuned for the robot embodiment to synthesize motion videos, which are then screened by an LLM quality filter. (ii) Each video that passes the filter is lifted into a 3D reference trajectory in the robot's configuration space by combining 2D keypoint detection with a URDF-anchored optimization; the recovered trajectory is re-rendered under the original camera viewpoint and retained only if its CLIP similarity to the source video exceeds a threshold and its reprojection error remains within geometric bounds, screening out extraction failures before any policy compute is spent. (iii) For each retained trajectory we train a PPO tracking policy following the BeyondMimic~\citep{liao2025beyondmimic} setup; trajectories whose tracking error exceeds a feasibility threshold are discarded, and the surviving policies are transferred to the real Go2. Because the generation, extraction, and training stages are automated, the resulting motion repertoire scales directly with compute.

\subsection{Identity-Consistent Video Generation}
\label{sec:wan_refinement}
For the extraction stage in Section~\ref{sec:trajectory} to lift these clips into 3D motions, the videos must satisfy three requirements: the robot's identity stays stable across every frame, the camera is fixed and known, and the background is stable. We fine-tune Wan2.2~\citep{wan2025} in a first-frame I2V setup, where each clip is conditioned on a canonical reference image $I_{\text{ref}}$ of the Go2 plus a Gemini~\citep{geminiteam2025geminifamilyhighlycapable}-generated motion prompt. The reference image fixes the camera and the curated training set fixes the background, so the last two requirements are met by construction. What the I2V setup does not handle is the first requirement: frame-to-frame identity stability. The flow-matching loss supervises each frame independently in latent space, with no constraint tying the identity of frame $T$ to that of frame $0$. The result is \emph{identity drift}, where the robot's body deforms non-rigidly and its appearance shifts across the clip, which in turn breaks downstream 3D extraction. We close this gap with a pixel-level identity consistency loss $\mathcal{L}_{\text{IC}}$ layered on top of $\mathcal{L}_{\text{FM}}$.

\textbf{Appearance bank}: We use the DINOv2~\citep{oquab2024dinov2learningrobustvisual} CLS embedding of each frame as its identity descriptor. The bank itself is built automatically from the training data rather than hand-picked. We pool every frame of every clip in the fine-tuning set, encode each through a frozen DINOv2 to obtain its CLS embedding, and then run a greedy coverage-set search over the resulting feature pool. Starting from an empty bank, at each step we add the frame that covers the largest number of still-uncovered frames within a cosine threshold $\tau$, and stop as soon as every pooled frame has at least one bank entry within $\tau$ in cosine similarity. With $\tau = 0.8$ this procedure converges in $N$ steps and produces a bank
\begin{equation*}
  \mathcal{B} = \{\, f_{\text{ref}}^{(j)} \,\}_{j=1}^{N}, \qquad f_{\text{ref}}^{(j)} = \text{DINOv2}\bigl(I_{\text{ref}}^{(j)}\bigr),
\end{equation*}
where each $I_{\text{ref}}^{(j)}$ is one of the selected training frames. By construction the bank is the smallest set of training-distribution snapshots that covers the entire fine-tuning pool within $\tau$; any in-distribution generated frame therefore has, by the same coverage guarantee, at least one bank entry within $\tau$ in feature space. We compute the bank once before training and freeze it.

\textbf{Nearest-reference hinge}: At each training step we reconstruct the predicted clean latent from the model's velocity output,
\begin{equation*}
  \hat{x}_0 = x_t - \sigma_t\, \hat{v}_\theta,
\end{equation*}
decode it through the frozen VAE into a candidate clean video $\hat{V}$, and subsample $T$ frames. Each frame goes through the same frozen DINOv2 to yield a CLS feature $f_t$, which we compare against every bank entry by cosine similarity. We keep only the nearest reference per frame and apply a hinge with margin $m_{\text{id}}$:
\begin{equation}
\label{eq:identity_loss}
  \mathcal{L}_{\text{IC}} = \frac{1}{T} \sum_{t=1}^{T} \max\!\left(0,\ m_{\text{id}} - \max_{j \in [N]} \cos\bigl(f_t,\, f_{\text{ref}}^{(j)}\bigr)\right).
\end{equation}
Because the maximum is taken over all bank entries, a frame only needs to be close to whichever canonical view it most resembles, so the natural feature shift caused by motion itself is not mistaken for identity drift. The margin $m_{\text{id}}$ absorbs the constant cosine offset that VAE decoding introduces, and is set from a one-off calibration pass on the baseline generator at a value just below the $\max_j \cos$ that non-degenerate generations typically produce.

We train a LoRA~\citep{hu2021loralowrankadaptationlarge} adapter on Wan2.2~\citep{wan2025} with the per-step objective
\begin{equation}
\label{eq:total_loss}
  \mathcal{L}_{\text{total}} = \mathcal{L}_{\text{FM}} + \lambda\, \mathcal{L}_{\text{IC}},
\end{equation}
where $\mathcal{L}_{\text{FM}}$ is the standard flow-matching loss, $\lambda$ is the identity weight. We attach $\mathcal{L}_{\text{IC}}$ only to Wan2.2's low-noise expert, where the predicted clean video $\hat{V}$ is reliable enough for DINOv2 to score perceptually; the high-noise expert continues to train under $\mathcal{L}_{\text{FM}}$ alone.

\subsection{From Video to 3D Reference Trajectory}
\label{sec:trajectory}

We convert each generated video into a 3D reference trajectory $\{\mathbf{s}_t\}_{t=1}^T$ in the robot's configuration space, and discard clips whose recovered motion does not faithfully reflect what the generator produced. Although recovering 3D pose from monocular video is ill-posed in general (depth, scale, and camera pose are all entangled with the unknown joint configuration), the I2V setting introduced in Section~\ref{sec:wan_refinement} removes these ambiguities by construction: the camera intrinsics and extrinsics are known and shared across every frame of every clip, and the frame-0 pose is exactly the URDF's canonical (default standing) state. With both the camera and the initial pose fixed, recovering the rest of the trajectory reduces to a temporally constrained kinematic fitting problem over per-frame joint angles and root motion.

\paragraph{Trajectory recovery.} For each frame we predict 2D positions of $K$ predefined body landmarks on the Go2 using a ViTPose~\citep{xu2022vitposesimplevisiontransformer} model fine-tuned on rendered images of the robot performing a set of basic motions. We then solve for the per-frame state $\mathbf{s}_t = (\mathbf{p}_t, \boldsymbol{\phi}_t, \boldsymbol{\theta}_t)$, comprising the root position $\mathbf{p}_t \in \mathbb{R}^3$, root Euler angles $\boldsymbol{\phi}_t \in \mathbb{R}^3$, and 12 actuated joint angles $\boldsymbol{\theta}_t \in \mathbb{R}^{12}$, with frame 0 initialized to the canonical URDF state. The objective is dominated by a 2D reprojection error that matches each URDF-projected keypoint to its detected 2D location,
\begin{equation} \label{eq:reproj}
L_{\text{reproj}}^{(t)} = \sum_{k} \| \Pi(\text{FK}(\mathbf{s}_t)_k) - \mathbf{p}^{(t)}_{2D,k} \|_2^2,
\end{equation}
where $\Pi$ is the projection under the fixed camera intrinsics and $\text{FK}$ is the URDF forward kinematics chain. We complement this with a temporal smoothness penalty on consecutive states and with foot-contact constraints that anchor any foot detected as in stance to the ground plane and to its previous-frame 3D position, preventing the recovered motion from drifting or sliding during contacts. For downstream storage and policy training, we convert the recovered Euler angles $\boldsymbol{\phi}_t$ to unit quaternions to avoid singularities and enable continuous interpolation, yielding the 19-D per-frame state vectors used in Section~\ref{sec:dataset}.           
\begin{figure}[t]
    \centering
    \includegraphics[width=\textwidth]{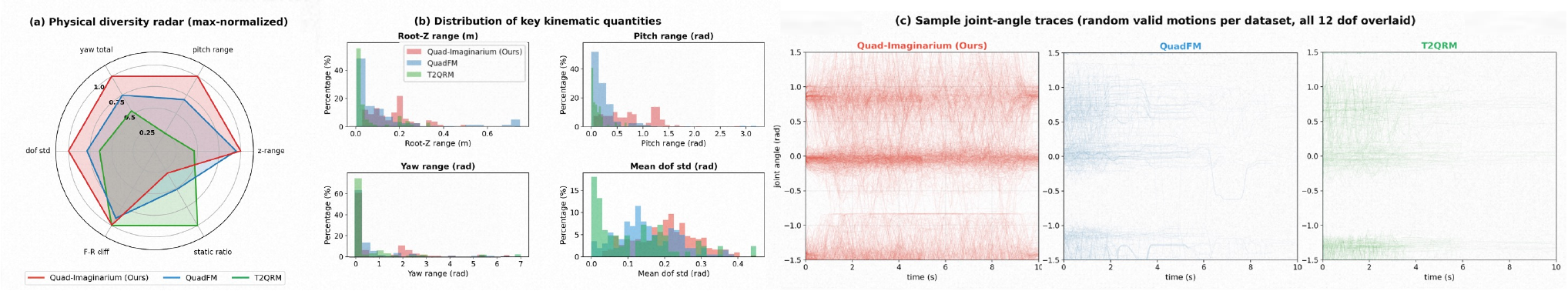}
    \caption{\textbf{Quad-Imaginarium} dominates both motion-capture-derived baselines on the majority of axes \textbf{(a)}, with histograms shifted toward higher values and heavier tails \textbf{(b)} and dense joint-angle coverage across the full $[-1.5, 1.5]$\,rad range \textbf{(c)}.}
    \label{fig:dataset}
\end{figure}

\paragraph{Multi-stage quality filtering.} The formulation above makes recovery tractable but does not guarantee success on every clip. We apply three sequential gates to discard unfaithful or infeasible trajectories.

\emph{CLIP semantic gate.} We re-render each recovered trajectory in MuJoCo under the same camera viewpoint used at generation time, subsample one frame every four from both the re-rendered clip and the original generated video, and compute the mean cosine similarity of CLIP image embeddings across all sampled frame pairs. Trajectories whose mean similarity falls below a threshold are discarded.

\emph{Geometric gate.} Trajectories that pass the semantic gate are further screened by reprojection error: we require both the per-clip mean reprojection error to stay below 20\,px and the per-clip max-frame error to stay below 100\,px. The dual threshold ensures overall alignment while catching motions that are well-aligned on average but contain individual frames with severe keypoint drift.

\emph{Tracking-error gate.} Even geometrically faithful trajectories may be physically infeasible. For each trajectory that passes the geometric gate, we train a PPO tracking policy and run a full-length episode in simulation. Trajectories on which the policy triggers a termination condition are discarded, retaining only motions that are physically executable.


\section{Quad-Imaginarium}
\label{sec:dataset}
We release \textbf{Quad-Imaginarium}, an open-source dataset of 7{,}488 quadruped motion clips at 24\,fps, totaling 18.5 hours, produced by the pipeline of Section~\ref{sec:overview}. Unlike prior quadruped motion datasets that derive content from real animal motion capture~\citep{peng2020imitateanimals,AcinoSet}, teleoperation, or artist authoring, every clip in Quad-Imaginarium is synthesized end-to-end---human effort enters only at the correction stage. Each clip is stored in the Go2 configuration space as per-frame 19-D state vectors (3-D root position, 4-D unit quaternion, 12 joint angles), and is accompanied by two complementary language annotations: a fine-grained \emph{description} of the motion, initially proposed by Gemini from the generated clip and then verified by human annotators, and a \emph{command instruction} corrected separately by human reviewers to reflect how a user would naturally invoke the same motion through speech. 
Because the motion generation and extraction pipeline is fully automated, the dataset can be continuously expanded by generating more prompts and allocating additional compute, with no marginal cost beyond GPU hours.

\textbf{Physical Diversity Analysis.} We compare Quad-Imaginarium against QuadFM~\citep{gao2026quadfmfoundationaltextdrivenquadruped} and T2QRM~\citep{10.1145/3696409.3700230} on six per-motion kinematic features (Figure~\ref{fig:dataset}): yaw total, pitch range, root-Z range, static ratio, front-rear leg motion asymmetry (F-R diff), and mean joint-angle standard deviation.

Quad-Imaginarium achieves the highest score on the majority of axes, with the largest margins on pitch range ---the axes that correspond to expressive, non-locomotion behaviors systematically missing from motion-capture-derived datasets. Its low static ratio further confirms that the motions are predominantly dynamic rather than static poses. This breadth of kinematic coverage is what underwrites the motion repertoire deployed on real hardware in Section~\ref{sec:experiments}: rather than reproducing a small set of well-rehearsed gaits, the policies trained on Quad-Imaginarium successfully execute the full spectrum of expressive behaviors the dataset spans.

\section{Experiments}
\label{sec:experiments}

\subsection{Video Generation Quality}
\label{sec:videogen_eval}

\textbf{Compared methods.} We evaluate three configurations of the Wan2.2~\citep{wan2025} I2V backbone: \emph{Wan-Base} (off-the-shelf, no adaptation), \emph{Wan-FT} (LoRA~\citep{hu2021loralowrankadaptationlarge} fine-tuned on curated robot videos under the flow-matching loss alone), and \emph{Wan-FT + $\mathcal{L}_{\text{IC}}$ (Ours)} (additionally trained with our Identity Consistency Loss, $\lambda{=}0.5$). All three share identical inference settings and are conditioned on the same 182 motion prompts with a unified Go2 reference image as the first frame.

\textbf{Metrics.} At the distribution level we report FID~\citep{chong2020effectivelyunbiasedfidinception} (InceptionV3 frame features) and FVD~\citep{unterthiner2019accurategenerativemodelsvideo} (R3D-18 spatiotemporal features). For per-clip semantics, Gemini-2.5-Pro~\citep{comanici2025gemini25pushingfrontier} scores each video on a 1--10 scale across Identity Consistency (IC), Naturalness (Nat), and Text-Action Alignment (Align). We additionally conducted a human survey on the same three dimensions, whose ratings positively correlate with the VLM scores (details in Appendix), indicating that Gemini-2.5-Pro's judgments faithfully reflect human perception.

\begin{figure}[t]
    \centering
    \includegraphics[width=\textwidth]{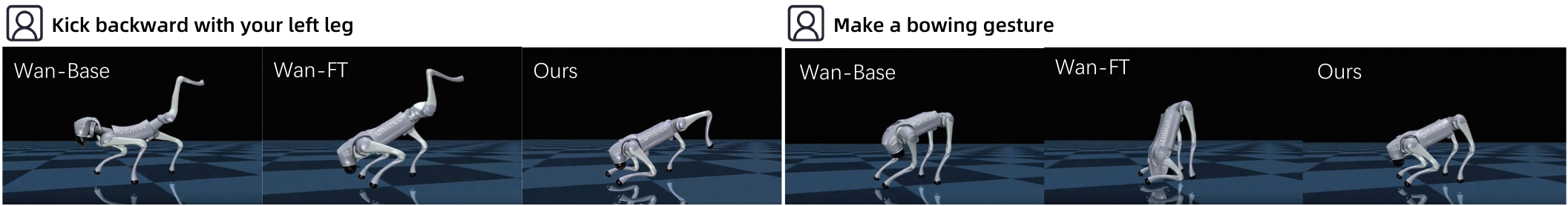}
    \caption{Qualitative comparison. Wan-Base shows severe body melting; Wan-FT still has occasional leg distortion; Wan-FT + $\mathcal{L}_{\text{IC}}$ maintains a clean rigid-body silhouette.}
    \label{fig:qualitative}
\end{figure}

Domain fine-tuning is the dominant driver of visual fidelity, with Wan-FT cutting both FID and FVD roughly in half relative to Wan-Base. Adding $\mathcal{L}_{\text{IC}}$ yields the best score on every axis and is the only configuration in which all five metrics improve together; the per-clip semantic deltas are small because the VLM scores are close to saturation, and the human survey shows a consistent ranking that supports this conclusion. The qualitative effect of $\mathcal{L}_{\text{IC}}$ is pronounced (Figure~\ref{fig:qualitative}).

\begin{table}[t]
\centering
\caption{Video generation quality across 182 test prompts.}
\label{tab:main_results}
\begin{tabular}{lccccc}
\toprule
Method & FID $\downarrow$ & FVD $\downarrow$ & IC $\uparrow$ & Nat $\uparrow$ & Align $\uparrow$ \\
\midrule
Wan-Base & 66.65 & 21.50 & 8.02 & 6.48 & 7.37 \\
Wan-FT & 27.95 & 10.63 & 8.30 & 6.67 & 7.48 \\
Wan-FT + $\mathcal{L}_{\text{IC}}$ (Ours) & \textbf{26.98} & \textbf{10.14} & \textbf{8.35} & \textbf{6.68} & \textbf{7.51} \\
\bottomrule
\end{tabular}
\end{table}




\subsection{Trajectory Extraction}
\label{sec:extraction_eval}

Extracted trajectories pass through two sequential gates: a CLIP~\citep{radford2021learningtransferablevisualmodels} semantic gate (mean per-clip CLIP cosine similarity between the re-rendered MuJoCo~\citep{6386109} trajectory and the source video $\geq 0.85$) and a geometric gate (per-clip mean reprojection error $< 20$\,px \emph{and} per-clip max-frame error $< 100$\,px). \textbf{97.0\% of candidates pass the semantic gate} (mean CLIP similarity 0.912, mean per-clip reprojection error 10.96\,px); of those, \textbf{70.2\% pass the geometric gate}, giving a \textbf{total retention rate of 68.1\%}. the geometric gate is the primary bottleneck. This conservative threshold is a deliberate design choice---over-generating and discarding is safer than relaxing thresholds and propagating extraction artifacts into policy training.


\subsection{Simulation Tracking}
\label{sec:tracking_eval}

For each of the 7{,}488 retained trajectories we train a PPO tracking policy following the BeyondMimic~\citep{liao2025beyondmimic} setup, counting an episode as successful if it completes the full reference motion without triggering any termination condition. \textbf{97.6\% of trajectories satisfy this criterion}; mean body-link position error (MBPE) is 40.6\,mm, mean root position error (MRPE) is 3.8\,cm, mean root orientation error (MROE) is 2.7\textdegree, and mean per-joint angle error (MPJPE) is 3.4\textdegree.


\subsection{Real Robot Deployment}
\label{sec:real_robot_eval}
\begin{figure}[t]
    \centering
    \includegraphics[width=\textwidth]{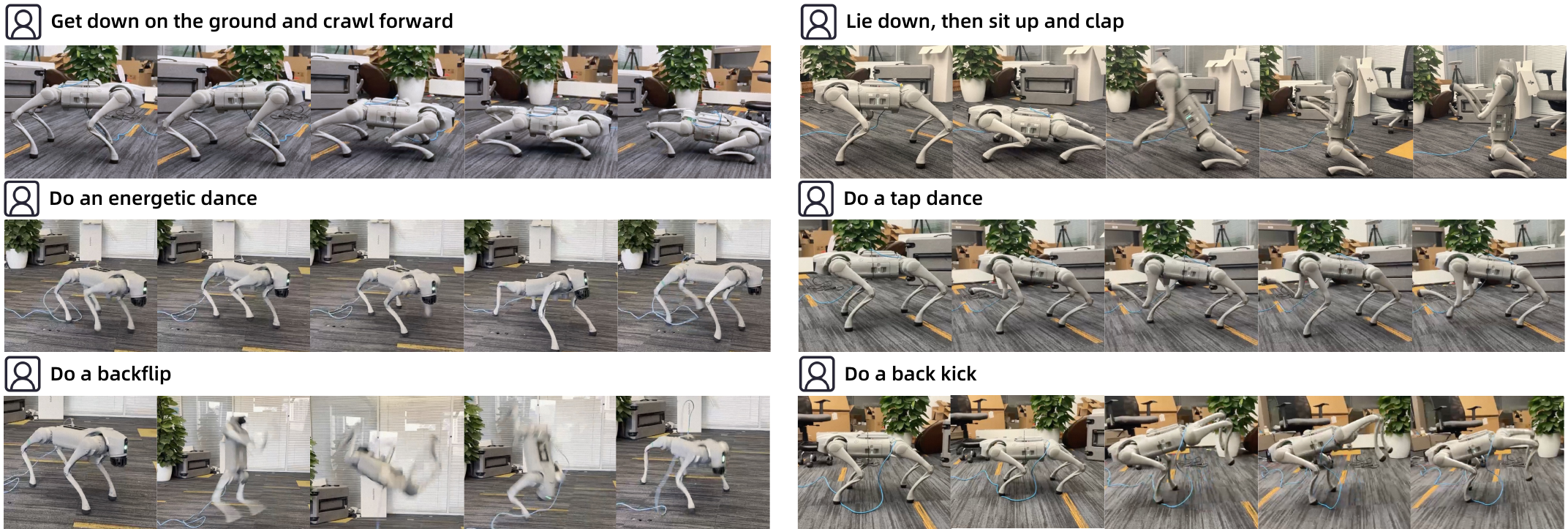}
    \caption{Representative real-robot executions of expressive motions from Quad-Imaginarium.}
    \label{fig:real_robot}
\end{figure}

We randomly sample 392 motions from Quad-Imaginarium, covering the full kinematic diversity of the dataset, and execute each $K{=}5$ times on a Unitree Go2 over flat indoor flooring without external perturbations. A trial is counted as successful if the robot completes the full motion without falling or requiring manual intervention. The pipeline achieves a \textbf{96.7\% deployment success rate} (Figure~\ref{fig:real_robot}). Combined with the 97.6\% simulation success rate reported in Section~\ref{sec:tracking_eval}, these results validate the full pipeline end-to-end. A detailed analysis of failure cases is provided in the Appendix.



\section{Limitations}
\label{sec:limits}
Uni-Mo's data engine inherits a well-known property of current video diffusion models: semantic control is imprecise, so a given prompt typically requires several generation attempts before a clip faithfully executes the requested motion. Our pipeline absorbs this by oversampling per prompt and discarding clips that fail manual prompt-video alignment verification. Since this verification is currently human rather than automatic, the marginal cost of expanding the dataset is dominated by curation labor rather than by generation alone. A second limitation comes from the assumption that the robot remains fully visible across every frame. Once the robot moves partially or fully out of the fixed camera view, the 2D keypoint observations required by our kinematic fitting become incomplete or unavailable, leaving the 3D trajectory unrecoverable; affected clips are silently discarded by the pipeline. As a result, Quad-Imaginarium is biased toward in-place expressive behaviors; extending the pipeline to large-translation locomotion would require either tracking cameras or a multi-view setup, neither of which we explored here.       

\section{Conclusion}
\label{sec:conclusion}
We presented Uni-Mo, an automated pipeline that turns natural-language prompts into deployable expressive motions for quadruped robots, removing the animal-in-the-loop assumption that has constrained prior work. Our Identity Consistency Loss closes the identity drift gap that otherwise prevents general-purpose video diffusion models from being used as a robot data engine. Policies trained on the released Quad-Imaginarium dataset achieve a 96.7\% deployment success rate on a real Unitree Go2 across 392 randomly sampled motions and 97.6\% success across the full 7{,}488-motion set in simulation. Expressive quadruped motion acquisition thus becomes a problem that scales with generation compute rather than with the cooperation of real animals.




\bibliography{example}  

@article{smpl,
author = {Loper, Matthew and Mahmood, Naureen and Romero, Javier and Pons-Moll, Gerard and Black, Michael J.},
title = {SMPL: a skinned multi-person linear model},
year = {2015},
issue_date = {November 2015},
publisher = {Association for Computing Machinery},
address = {New York, NY, USA},
volume = {34},
number = {6},
issn = {0730-0301},
url = {https://doi.org/10.1145/2816795.2818013},
doi = {10.1145/2816795.2818013},
month = nov,
articleno = {248},
numpages = {16}
}

@misc{wang2025dogmolargescalemultiviewrgbd,
      title={DogMo: A Large-Scale Multi-View RGB-D Dataset for 4D Canine Motion Recovery}, 
      author={Zan Wang and Siyu Chen and Luya Mo and Xinfeng Gao and Yuxin Shen and Lebin Ding and Wei Liang},
      year={2025},
      eprint={2510.24117},
      archivePrefix={arXiv},
      primaryClass={cs.CV},
      url={https://arxiv.org/abs/2510.24117}, 
}

@INPROCEEDINGS{AcinoSet,
  author={Joska, Daniel and Clark, Liam and Muramatsu, Naoya and Jericevich, Ricardo and Nicolls, Fred and Mathis, Alexander and Mathis, Mackenzie W. and Patel, Amir},
  booktitle={2021 IEEE International Conference on Robotics and Automation (ICRA)}, 
  title={AcinoSet: A 3D Pose Estimation Dataset and Baseline Models for Cheetahs in the Wild}, 
  year={2021},
  volume={},
  number={},
  pages={13901-13908},
  doi={10.1109/ICRA48506.2021.9561338}}

@conference{AMASS,
  title           = {{AMASS}: Archive of Motion Capture as Surface Shapes},
  author          = {Mahmood, Naureen and Ghorbani, Nima and Troje, Nikolaus F. and Pons-Moll, Gerard and Black, Michael J.},
  booktitle       = {International Conference on Computer Vision},
  pages           = {5442--5451},
  month           = oct,
  year            = {2019},
  month_numeric   = {10}
}

@misc{AMASS_CMU,
  title           = {{CMU MoCap Dataset}},
  author          = {{Carnegie Mellon University}},
  url             = {http://mocap.cs.cmu.edu}
}

@misc{liao2025beyondmimic,
      title={BeyondMimic: From Motion Tracking to Versatile Humanoid Control via Guided Diffusion}, 
      author={Qiayuan Liao and Takara E. Truong and Xiaoyu Huang and Yuman Gao and Guy Tevet and Koushil Sreenath and C. Karen Liu},
      year={2025},
      eprint={2508.08241},
      archivePrefix={arXiv},
      primaryClass={cs.RO},
      url={https://arxiv.org/abs/2508.08241}, 
}

@misc{mao2025physworld,
      title={Robot Learning from a Physical World Model}, 
      author={Jiageng Mao and Sicheng He and Hao-Ning Wu and Yang You and Shuyang Sun and Zhicheng Wang and Yanan Bao and Huizhong Chen and Leonidas Guibas and Vitor Guizilini and Howard Zhou and Yue Wang},
      year={2025},
      eprint={2511.07416},
      archivePrefix={arXiv},
      primaryClass={cs.RO},
      url={https://arxiv.org/abs/2511.07416}, 
}

@misc{ye2025gen2real,
      title={Gen2Real: Towards Demo-Free Dexterous Manipulation by Harnessing Generated Video}, 
      author={Kai Ye and Yuhang Wu and Shuyuan Hu and Junliang Li and Meng Liu and Yongquan Chen and Rui Huang},
      year={2025},
      eprint={2509.14178},
      archivePrefix={arXiv},
      primaryClass={cs.RO},
      url={https://arxiv.org/abs/2509.14178}, 
}

@misc{ni2025genmimic,
      title={From Generated Human Videos to Physically Plausible Robot Trajectories}, 
      author={James Ni and Zekai Wang and Wei Lin and Amir Bar and Yann LeCun and Trevor Darrell and Jitendra Malik and Roei Herzig},
      year={2025},
      eprint={2512.05094},
      archivePrefix={arXiv},
      primaryClass={cs.RO},
      url={https://arxiv.org/abs/2512.05094}, 
}

@misc{wu2025quadreamer,
      title={QuaDreamer: Controllable Panoramic Video Generation for Quadruped Robots}, 
      author={Sheng Wu and Fei Teng and Hao Shi and Qi Jiang and Kai Luo and Kaiwei Wang and Kailun Yang},
      year={2025},
      eprint={2508.02512},
      archivePrefix={arXiv},
      primaryClass={cs.RO},
      url={https://arxiv.org/abs/2508.02512}, 
}

@misc{tang2025trajskill,
      title={Trajectory Conditioned Cross-embodiment Skill Transfer}, 
      author={YuHang Tang and Yixuan Lou and Pengfei Han and Haoming Song and Xinyi Ye and Dong Wang and Bin Zhao},
      year={2025},
      eprint={2510.07773},
      archivePrefix={arXiv},
      primaryClass={cs.RO},
      url={https://arxiv.org/abs/2510.07773}, 
}

@misc{li2025novaflow,
      title={NovaFlow: Zero-Shot Manipulation via Actionable Flow from Generated Videos}, 
      author={Hongyu Li and Lingfeng Sun and Yafei Hu and Duy Ta and Jennifer Barry and George Konidaris and Jiahui Fu},
      year={2025},
      eprint={2510.08568},
      archivePrefix={arXiv},
      primaryClass={cs.RO},
      url={https://arxiv.org/abs/2510.08568}, 
}

@article{albaba2025nil,
  title={NIL: No-data Imitation Learning by Leveraging Pre-trained Video Diffusion Models},
  author={Albaba, Mert and Li, Chen and Diomataris, Markos and Taheri, Omid and Krause, Andreas and Black, Michael J},
  journal={arXiv preprint arXiv:2503.10626},
  year={2025}
}

@article{yoon2024stmr,
  title={Spatio-Temporal Motion Retargeting for Quadruped Robots},
  author={Yoon, Taerim and Kang, Dongho and Kim, Seunghun and Ahn, Minsung and Coros, Stelian and Choi, Sungjoon},
  journal={IEEE Transactions on Robotics},
  volume={41},
  pages={5471--5490},
  year={2024},
  doi={10.1109/TRO.2025.3600123}
}

@inproceedings{yang2023vim,
  title={Generalized Animal Imitator: Agile Locomotion with Versatile Motion Prior},
  author={Yang, Ruihan and others},
  booktitle={Conference on Robot Learning},
  pages={4631--4650},
  year={2023}
}

@inproceedings{li2023terrainadaptive,
  title={Learning Terrain-Adaptive Locomotion with Agile Behaviors by Imitating Animals},
  author={Li, Tingguang and others},
  booktitle={2023 IEEE/RSJ International Conference on Intelligent Robots and Systems (IROS)},
  pages={339--345},
  year={2023},
  doi={10.1109/IROS55552.2023.10342271}
}

@inproceedings{allshire2025videomimic,
  title={Visual Imitation Enables Contextual Humanoid Control},
  author={Allshire, Arthur and Choi, Hongsuk and Zhang, Junyi and McAllister, David and Zhang, Anthony and Kim, Chung Min and Darrell, Trevor and Abbeel, Pieter and Malik, Jitendra and Kanazawa, Angjoo},
  booktitle={9th Conference on Robot Learning (CoRL)},
  year={2025}
}

@inproceedings{peng2020imitateanimals,
  title={Learning Agile Robotic Locomotion Skills by Imitating Animals},
  author={Peng, Xue Bin and Coumans, Erwin and Zhang, Tingnan and Lee, Tsang-Wei and Tan, Jie and Levine, Sergey},
  booktitle={Robotics: Science and Systems},
  year={2020},
  doi={10.15607/rss.2020.xvi.064}
}

@article{peng2021amp,
  title={AMP: Adversarial Motion Priors for Stylized Physics-Based Character Control},
  author={Peng, Xue Bin and Ma, Ze and Abbeel, Pieter and Levine, Sergey and Kanazawa, Angjoo},
  journal={ACM Transactions on Graphics (TOG)},
  volume={40},
  number={4},
  pages={1--20},
  year={2021},
  doi={10.1145/3450626.3459670}
}

@inproceedings{escontrela2022amp_robots,
  title={Adversarial Motion Priors Make Good Substitutes for Complex Reward Functions},
  author={Escontrela, Alejandro and others},
  booktitle={2022 IEEE/RSJ International Conference on Intelligent Robots and Systems (IROS)},
  pages={25--32},
  year={2022},
  doi={10.1109/IROS47612.2022.9981973}
}

@inproceedings{bohez2022imitate,
  title={Imitate and Repurpose: Learning Reusable Robot Movement Skills From Human and Animal Behaviors},
  author={Bohez, Steven and others},
  booktitle={arXiv preprint arXiv:2203.17138},
  year={2022}
}

@article{zhang2023slomo,
  title={SLoMo: A General System for Legged Robot Motion Imitation From Casual Videos},
  author={Zhang, John Z and others},
  journal={IEEE Robotics and Automation Letters},
  volume={8},
  pages={7154--7161},
  year={2023},
  doi={10.1109/LRA.2023.3313937}
}

@misc{chanesane2024rlwav,
      title={Reinforcement Learning from Wild Animal Videos}, 
      author={Elliot Chane-Sane and Constant Roux and Olivier Stasse and Nicolas Mansard},
      year={2024},
      eprint={2412.04273},
      archivePrefix={arXiv},
      primaryClass={cs.RO},
      url={https://arxiv.org/abs/2412.04273}, 
}

@article{zhao2025aggressive,
  title={Learning aggressive animal locomotion skills for quadrupedal robots solely from monocular videos},
  author={Zhao, Liu and Luo, Zeren and Han, Yimin and Zhang, Jiahui and Chen, Yuanhao and Liu, Yunhui and Lu, Peng},
  journal={npj Robotics},
  volume={3},
  number={1},
  pages={32},
  year={2025},
  publisher={Nature Publishing Group UK London}
}

@article{huang2024diffuseloco,
  title={DiffuseLoco: Real-Time Legged Locomotion Control with Diffusion from Offline Datasets},
  author={Huang, Xiaoyu and others},
  journal={arXiv preprint arXiv:2404.19264},
  year={2024}
}

@misc{chen2025inbetween,
      title={In-between Motion Generation Based Multi-Style Quadruped Robot Locomotion}, 
      author={Yuanhao Chen and Liu Zhao and Ji Ma and Peng Lu},
      year={2025},
      eprint={2507.23053},
      archivePrefix={arXiv},
      primaryClass={cs.RO},
      url={https://arxiv.org/abs/2507.23053}, 
}

@article{yu2021ap10k,
  title={AP-10K: A Benchmark for Animal Pose Estimation in the Wild},
  author={Yu, Hang and Xu, Yufei and Zhang, Jing and Zhao, Wei and Guan, Ziyu and Tao, Dacheng},
  journal={arXiv preprint arXiv:2108.12617},
  year={2021}
}

@inproceedings{10.1145/3696409.3700230,
author = {Wang, Minghui and Wang, Zixu and Xu, Hongbin and Hu, Kun and Wang, Zhiyong and Kang, Wenxiong},
title = {T2QRM: Text-Driven Quadruped Robot Motion Generation},
year = {2024},
isbn = {9798400712739},
publisher = {Association for Computing Machinery},
address = {New York, NY, USA},
url = {https://doi.org/10.1145/3696409.3700230},
doi = {10.1145/3696409.3700230},
booktitle = {Proceedings of the 6th ACM International Conference on Multimedia in Asia},
articleno = {69},
numpages = {7},
location = {
},
series = {MMAsia '24}
}

@misc{gao2026quadfmfoundationaltextdrivenquadruped,
      title={QuadFM: Foundational Text-Driven Quadruped Motion Dataset for Generation and Control}, 
      author={Li Gao and Fuzhi Yang and Jianhui Chen and Liu Liu and Yao Zheng and Yang Cai and Ziqiao Li},
      year={2026},
      eprint={2603.24021},
      archivePrefix={arXiv},
      primaryClass={cs.RO},
      url={https://arxiv.org/abs/2603.24021}, 
}

@misc{mou2025dimodiverse3dmotion,
      title={DIMO: Diverse 3D Motion Generation for Arbitrary Objects}, 
      author={Linzhan Mou and Jiahui Lei and Chen Wang and Lingjie Liu and Kostas Daniilidis},
      year={2025},
      eprint={2511.07409},
      archivePrefix={arXiv},
      primaryClass={cs.CV},
      url={https://arxiv.org/abs/2511.07409}, 
}

@INPROCEEDINGS{10610200,
  author={Cheng, Xuxin and Shi, Kexin and Agarwal, Ananye and Pathak, Deepak},
  booktitle={2024 IEEE International Conference on Robotics and Automation (ICRA)}, 
  title={Extreme Parkour with Legged Robots}, 
  year={2024},
  volume={},
  number={},
  pages={11443-11450},
  doi={10.1109/ICRA57147.2024.10610200}}

@inproceedings{inproceedings,
author = {Kumar, Ashish and Fu, Zipeng and Pathak, Deepak and Malik, Jitendra},
year = {2021},
month = {07},
pages = {},
title = {RMA: Rapid Motor Adaptation for Legged Robots},
doi = {10.15607/RSS.2021.XVII.011}
}

@misc{cheng2024expressivewholebodycontrolhumanoid,
      title={Expressive Whole-Body Control for Humanoid Robots}, 
      author={Xuxin Cheng and Yandong Ji and Junming Chen and Ruihan Yang and Ge Yang and Xiaolong Wang},
      year={2024},
      eprint={2402.16796},
      archivePrefix={arXiv},
      primaryClass={cs.RO},
      url={https://arxiv.org/abs/2402.16796}, 
}

@misc{fu2024humanplushumanoidshadowingimitation,
      title={HumanPlus: Humanoid Shadowing and Imitation from Humans}, 
      author={Zipeng Fu and Qingqing Zhao and Qi Wu and Gordon Wetzstein and Chelsea Finn},
      year={2024},
      eprint={2406.10454},
      archivePrefix={arXiv},
      primaryClass={cs.RO},
      url={https://arxiv.org/abs/2406.10454}, 
}

@article{10.1145/3528223.3530110,
author = {Peng, Xue Bin and Guo, Yunrong and Halper, Lina and Levine, Sergey and Fidler, Sanja},
title = {ASE: large-scale reusable adversarial skill embeddings for physically simulated characters},
year = {2022},
issue_date = {July 2022},
publisher = {Association for Computing Machinery},
address = {New York, NY, USA},
volume = {41},
number = {4},
issn = {0730-0301},
url = {https://doi.org/10.1145/3528223.3530110},
doi = {10.1145/3528223.3530110},
journal = {ACM Trans. Graph.},
month = jul,
articleno = {94},
numpages = {17}
}

@article{rudin2021learning,
title={Learning to Walk in Minutes Using Massively Parallel Deep Reinforcement Learning},
author={Rudin, Nikita and Hoeller, David and Reist, Philipp and Hutter, Marco},
journal={arXiv preprint arXiv:2109.11978},
year={2021}
}

@article{haarnoja2024soccer,
title={Learning Agile Soccer Skills for a Bipedal Robot with Deep Reinforcement Learning},
author={Haarnoja, Tuomas and Moran, Ben and Lever, Guy and Huang, Sandy H. and others},
journal={Science Robotics},
volume={9},
number={89},
year={2024},
doi={10.1126/scirobotics.adi8022}
}

@article{ji2024exbody2,
title={ExBody2: Advanced Expressive Humanoid Whole-Body Control},
author={Ji, Mazeyu and Peng, Xuanbin and Liu, Fangchen and Li, Jialong and Yang, Ge and Cheng, Xuxin and Wang, Xiaolong},
journal={arXiv preprint arXiv:2412.13196},
year={2024}
}

@article{he2024hover,
title={HOVER: Versatile Neural Whole-Body Controller for Humanoid Robots},
author={He, Tairan and Xiao, Wenli and Lin, Toru and Luo, Zhengyi and Xu, Zhenjia and Jiang, Zhenyu and Kautz, Jan and Liu, Changliu and Shi, Guanya and Wang, Xiaolong and Fan, Linxi and Zhu, Yuke},
journal={arXiv preprint arXiv:2410.21229},
year={2024}
}

@article{wococo2024,
title={WoCoCo: Learning Whole-Body Humanoid Control with Sequential Contacts},
author={Zhang, Chong and Xiao, Wenli and He, Tairan and Shi, Guanya},
journal={arXiv preprint arXiv:2406.06005},
year={2024}
}

@article{zhuang2024humanoidparkour,
title={Humanoid Parkour Learning},
author={Zhuang, Ziwen and Yao, Shenzhe and Zhao, Hang},
journal={arXiv preprint arXiv:2406.10759},
year={2024}
}

@article{chen2022mld,
title={Executing your Commands via Motion Diffusion in Latent Space},
author={Chen, Xin and Jiang, Biao and Liu, Wen and Huang, Zilong and Fu, Bin and Chen, Tao and Yu, Jingyi and Yu, Gang},
journal={arXiv preprint arXiv:2212.04048},
year={2023}
}

@article{wan2025,
title={Wan: Open and Advanced Large-Scale Video Generative Models},
author={{Wan Team} and Wang, Ang and Ai, Baole and Wen, Bin and Mao, Chaojie and others},
journal={arXiv preprint arXiv:2503.20314},
year={2025}
}

@article{kong2024hunyuanvideo,
title={HunyuanVideo: A Systematic Framework For Large Video Generative Models},
author={Kong, Weijie and Tian, Qi and Zhang, Zijian and Min, Rox and Dai, Zuozhuo and Zhou, Jin and others},
journal={arXiv preprint arXiv:2412.03603},
year={2024}
}

@misc{seedance2026seedance20advancingvideo,
      title={Seedance 2.0: Advancing Video Generation for World Complexity}, 
      author={{Team Seedance} and De Chen and Liyang Chen and Xin Chen and Ying Chen  and others},
      year={2026},
      eprint={2604.14148},
      archivePrefix={arXiv},
      primaryClass={cs.CV},
      url={https://arxiv.org/abs/2604.14148}, 
}

@misc{geminiteam2025geminifamilyhighlycapable,
      title={Gemini: A Family of Highly Capable Multimodal Models}, 
      author={{Gemini Team} and Rohan Anil and Sebastian Borgeaud and Jean-Baptiste Alayrac and others},
      year={2025},
      eprint={2312.11805},
      archivePrefix={arXiv},
      primaryClass={cs.CL},
      url={https://arxiv.org/abs/2312.11805}, 
}

@misc{oquab2024dinov2learningrobustvisual,
      title={DINOv2: Learning Robust Visual Features without Supervision}, 
      author={Maxime Oquab and Timothée Darcet and Théo Moutakanni and Huy Vo and Marc Szafraniec and Vasil Khalidov and Pierre Fernandez and Daniel Haziza and Francisco Massa and Alaaeldin El-Nouby and Mahmoud Assran and Nicolas Ballas and Wojciech Galuba and Russell Howes and Po-Yao Huang and Shang-Wen Li and Ishan Misra and Michael Rabbat and Vasu Sharma and Gabriel Synnaeve and Hu Xu and Hervé Jegou and Julien Mairal and Patrick Labatut and Armand Joulin and Piotr Bojanowski},
      year={2024},
      eprint={2304.07193},
      archivePrefix={arXiv},
      primaryClass={cs.CV},
      url={https://arxiv.org/abs/2304.07193}, 
}

@misc{hu2021loralowrankadaptationlarge,
      title={LoRA: Low-Rank Adaptation of Large Language Models}, 
      author={Edward J. Hu and Yelong Shen and Phillip Wallis and Zeyuan Allen-Zhu and Yuanzhi Li and Shean Wang and Lu Wang and Weizhu Chen},
      year={2021},
      eprint={2106.09685},
      archivePrefix={arXiv},
      primaryClass={cs.CL},
      url={https://arxiv.org/abs/2106.09685}, 
}

@misc{xu2022vitposesimplevisiontransformer,
      title={ViTPose: Simple Vision Transformer Baselines for Human Pose Estimation}, 
      author={Yufei Xu and Jing Zhang and Qiming Zhang and Dacheng Tao},
      year={2022},
      eprint={2204.12484},
      archivePrefix={arXiv},
      primaryClass={cs.CV},
      url={https://arxiv.org/abs/2204.12484}, 
}

@misc{comanici2025gemini25pushingfrontier,
      title={Gemini 2.5: Pushing the Frontier with Advanced Reasoning, Multimodality, Long Context, and Next Generation Agentic Capabilities}, 
      author={Gheorghe Comanici and Eric Bieber and Mike Schaekermann and Ice Pasupat and others},
      year={2025},
      eprint={2507.06261},
      archivePrefix={arXiv},
      primaryClass={cs.CL},
      url={https://arxiv.org/abs/2507.06261}, 
}

@misc{chong2020effectivelyunbiasedfidinception,
      title={Effectively Unbiased FID and Inception Score and where to find them}, 
      author={Min Jin Chong and David Forsyth},
      year={2020},
      eprint={1911.07023},
      archivePrefix={arXiv},
      primaryClass={cs.CV},
      url={https://arxiv.org/abs/1911.07023}, 
}

@misc{unterthiner2019accurategenerativemodelsvideo,
      title={Towards Accurate Generative Models of Video: A New Metric \& Challenges},
      author={Thomas Unterthiner and Sjoerd van Steenkiste and Karol Kurach and Raphael Marinier and Marcin Michalski and Sylvain Gelly},
      year={2019},
      eprint={1812.01717},
      archivePrefix={arXiv},
      primaryClass={cs.CV},
      url={https://arxiv.org/abs/1812.01717}, 
}

@misc{radford2021learningtransferablevisualmodels,
      title={Learning Transferable Visual Models From Natural Language Supervision}, 
      author={Alec Radford and Jong Wook Kim and Chris Hallacy and Aditya Ramesh and Gabriel Goh and Sandhini Agarwal and Girish Sastry and Amanda Askell and Pamela Mishkin and Jack Clark and Gretchen Krueger and Ilya Sutskever},
      year={2021},
      eprint={2103.00020},
      archivePrefix={arXiv},
      primaryClass={cs.CV},
      url={https://arxiv.org/abs/2103.00020}, 
}

@INPROCEEDINGS{6386109,
  author={Todorov, Emanuel and Erez, Tom and Tassa, Yuval},
  booktitle={2012 IEEE/RSJ International Conference on Intelligent Robots and Systems}, 
  title={MuJoCo: A physics engine for model-based control}, 
  year={2012},
  volume={},
  number={},
  pages={5026-5033},
  doi={10.1109/IROS.2012.6386109}}

@INPROCEEDINGS {8100069,
author = { Zuffi, Silvia and Kanazawa, Angjoo and Jacobs, David W. and Black, Michael J. },
booktitle = { 2017 IEEE Conference on Computer Vision and Pattern Recognition (CVPR) },
title = {{ 3D Menagerie: Modeling the 3D Shape and Pose of Animals }},
year = {2017},
volume = {},
ISSN = {1063-6919},
pages = {5524-5532},
doi = {10.1109/CVPR.2017.586},
url = {https://doi.ieeecomputersociety.org/10.1109/CVPR.2017.586},
publisher = {IEEE Computer Society},
address = {Los Alamitos, CA, USA},
month =Jul}

@misc{zakka2026mjlablightweightframeworkgpuaccelerated,
  title={mjlab: A Lightweight Framework for GPU-Accelerated Robot Learning},
  author={Kevin Zakka and Qiayuan Liao and Brent Yi and Louis Le Lay and Koushil Sreenath and Pieter Abbeel},
  year={2026},
  eprint={2601.22074},
  archivePrefix={arXiv},
  primaryClass={cs.RO},
  url={https://arxiv.org/abs/2601.22074},
}

\appendix

\section{Implementation Details}
\label{app:implementation}

\subsection{Video Generation Model Fine-tuning}
\label{app:video_gen}

We fine-tune Wan2.2-I2V-A14B~\citep{wan2025} using LoRA~\citep{hu2021loralowrankadaptationlarge} with rank 32, targeting the attention projection matrices and feed-forward layers of the low-noise diffusion expert. Training uses a learning rate of $1 \times 10^{-4}$ for 10 epochs with gradient checkpointing enabled. The model is trained on 56 NVIDIA H20 GPUs with a per-GPU batch size of 1 and gradient accumulation steps of 1, with checkpoints saved every 200 steps. The low-noise expert is trained on the timestep range $[0.358, 1.0]$, following Wan2.2's dual-expert architecture.

\paragraph{Training data.} The training set comprises 190 unique designer-choreographed motion sequences on the Unitree Go2 in MuJoCo~\citep{6386109}, each rendered from 10 distinct camera viewpoints, producing 1{,}900 video--prompt pairs in total. Each clip is 240 frames at 832$\times$480 resolution. The prompts are structured instructions that describe the robot's motion while explicitly requesting fixed camera pose, unchanged scene, and no subject deformation (e.g., ``Using the given image as the first and last frame, generate a video of the silver quadruped robot dog looking around curiously on a reflective floor, keeping camera pose unchanged, camera angle unchanged, scene unchanged, and no subject deformation'').

\paragraph{Identity Consistency Loss hyperparameters.} We set the identity loss weight $\lambda = 0.5$. The appearance bank is constructed with cosine threshold $\tau = 0.8$; we select $N = 20$ canonical frames via the greedy coverage-set procedure described in Section~\ref{app:ablation_bank}. At each training step we subsample $T$ frames from the decoded video for DINOv2~\citep{oquab2024dinov2learningrobustvisual} feature extraction.

\paragraph{Inference settings.} At inference time, all models generate at 832$\times$480 resolution, 24\,fps, 240 frames, conditioned on the same canonical Go2 reference image as the first frame.

\subsection{PPO Tracking Policy Training}
\label{app:ppo}

We train PPO tracking policies following the BeyondMimic~\citep{liao2025beyondmimic} framework, re-implemented in MuJoCo~\citep{6386109} with the MjLab~\citep{zakka2026mjlablightweightframeworkgpuaccelerated} infrastructure. Training uses 4{,}096 parallel environments on one NVIDIA 3090 GPU for up to 3{,}000 iterations per motion.

\paragraph{Network architecture.} Both actor and critic are multi-layer perceptrons with hidden dimensions $(512, 256, 128)$ and ELU activations. The actor outputs a Gaussian distribution (scalar standard deviation, initialized at 1.0) over 12-dimensional joint position actions. The critic receives a privileged observation that includes ground-truth body positions and orientations. Both networks use running observation normalization.

\paragraph{PPO hyperparameters.} We use a clipping parameter of $\epsilon = 0.2$, entropy coefficient $0.005$, value loss coefficient $1.0$ with clipped value loss, discount factor $\gamma = 0.99$, GAE parameter $\lambda_{\text{GAE}} = 0.95$, and adaptive learning rate starting at $1 \times 10^{-3}$ with a target KL divergence of $0.01$. Each update performs 5 learning epochs over 4 mini-batches, collecting 24 steps per environment before each update.

\paragraph{Action space.} The policy outputs residual joint position targets scaled by 0.25 and added to the default standing joint angles, yielding PD position commands at 50\,Hz (simulation timestep 5\,ms with decimation factor 4).


\paragraph{Reward function.} The reward is a weighted sum of tracking terms and regularization penalties, summarized in Table~\ref{tab:reward}.

\begin{table}[h]
\centering
\caption{Reward terms for the PPO tracking policy.}
\label{tab:reward}
\begin{tabular}{llc}
\toprule
Term & Description & Weight \\
\midrule
\multicolumn{3}{l}{\emph{Tracking rewards}} \\
\midrule
Global root position & $\|\mathbf{p}_{\text{ref}} - \mathbf{p}_{\text{robot}}\|^2$ & 0.5 \\
Global root orientation & Quaternion geodesic error & 0.8 \\
Relative body position & Mean link position error & 3.0 \\
Relative body orientation & Mean link orientation error & 1.5 \\
Body linear velocity & Mean link velocity error & 0.1 \\
Body angular velocity & Mean link angular velocity error & 0.3 \\
Anchor upward velocity & Upward velocity tracking & 1.5 \\
Anchor pitch rate & Pitch angular velocity tracking & 1.8 \\
\midrule
\multicolumn{3}{l}{\emph{Regularization penalties}} \\
\midrule
Action rate & $\|\mathbf{a}_t - \mathbf{a}_{t-1}\|^2$ & $-0.01$ \\
Joint limits & Soft penalty near joint bounds & $-10.0$ \\
Self-collision & Contact force exceeding threshold & $-10.0$ \\
\bottomrule
\end{tabular}
\end{table}

\paragraph{Termination conditions.} An episode terminates early if any of the following conditions is met: (i) the anchor (root) height deviates from the reference by more than 0.25\,m; (ii) the projected gravity direction of the anchor diverges from the reference by more than 0.8 (cosine); (iii) any end-effector (FL/FR/RL/RR calf) height deviates from the reference by more than 0.25\,m.

\section{Human Survey Details}
\label{app:human_survey}

To validate that the VLM-as-Judge scores faithfully reflect human perception, we conducted a human evaluation study.

\paragraph{Survey design.} We built a custom HTML questionnaire. Evaluators were presented with 20 video groups, each containing three anonymized videos (labeled Method~A, Method~B, Method~C) corresponding to Wan-Base, Wan-FT, and Wan-FT~+~$\mathcal{L}_{\text{IC}}$, respectively. The method--label mapping was not disclosed to evaluators. For each video, evaluators rated three dimensions on a 1--10 slider scale:
\begin{itemize}
    \item \textbf{Identity Consistency (IC):} Whether the robot maintains its rigid-body structure throughout the video. 10 = perfect, no deformation; 1 = severe melting/distortion.
    \item \textbf{Naturalness (Nat):} Whether the motion is physically plausible and fluid. 10 = completely natural; 1 = severely violates physics.
    \item \textbf{Text--Action Alignment (Align):} Whether the robot's actions match the text prompt. 10 = perfect match; 1 = completely unrelated.
\end{itemize}

\paragraph{Video selection.} The 20 video groups were randomly sampled from the 182-prompt test set.

\paragraph{Participants.} 20 evaluators participated, all with computer science or robotics backgrounds. Each evaluator completed all 20 groups independently, yielding 400 scored video triplets in total.

\paragraph{Results.} The mean scores are reported in Table~\ref{tab:human_survey}. The human ratings exhibit the same monotonic ordering as the VLM scores across all three dimensions: Wan-FT~+~$\mathcal{L}_{\text{IC}}$ $>$ Wan-FT $>$ Wan-Base.

\begin{table}[h]
\centering
\caption{Human evaluation results.}
\label{tab:human_survey}
\begin{tabular}{lccc}
\toprule
Method & IC $\uparrow$ & Nat $\uparrow$ & Align $\uparrow$ \\
\midrule
Wan-Base & 4.64 & 4.16 & 2.84 \\
Wan-FT & 7.04 & 5.96 & 4.51 \\
Wan-FT + $\mathcal{L}_{\text{IC}}$ (Ours) & \textbf{7.30} & \textbf{6.24} & \textbf{4.75} \\
\bottomrule
\end{tabular}
\end{table}

The consistent ordering between human and VLM scores confirms that Gemini-2.5-Pro's automatic judgments are a reliable proxy for human perception in this evaluation setting, validating the use of VLM-as-Judge for scalable evaluation.


\section{Ablation Studies}
\label{app:ablation}

\subsection{Identity Consistency Loss Weight $\lambda$}
\label{app:ablation_lambda}

We ablate the weight $\lambda$ of the Identity Consistency Loss $\mathcal{L}_{\text{IC}}$ by training LoRA adapters with $\lambda \in \{0, 0.1, 0.5\}$, where $\lambda = 0$ corresponds to Wan-FT (flow-matching loss only). All other hyperparameters are held constant, and results on the 182-prompt test set are reported in Table~\ref{tab:ablation_lambda}.

\begin{table}[h]
\centering
\caption{Ablation on Identity Consistency Loss weight $\lambda$.}
\label{tab:ablation_lambda}
\begin{tabular}{lccccc}
\toprule
$\lambda$ & FID $\downarrow$ & FVD $\downarrow$ & IC $\uparrow$ & Nat $\uparrow$ & Align $\uparrow$ \\
\midrule
0 & 27.95 & 10.63 & 8.30 & 6.67 & 7.48 \\
0.1 & 27.07 & 10.30 & 8.35 & 6.60 & 7.50 \\
0.5 & \textbf{26.98} & \textbf{10.14} & \textbf{8.35} & \textbf{6.68} & \textbf{7.51} \\
\bottomrule
\end{tabular}
\end{table}

Increasing $\lambda$ consistently improves both distribution-level metrics (FID, FVD) and the identity consistency score, without degrading naturalness or text alignment. The improvement from $\lambda = 0.1$ to $\lambda = 0.5$ is modest on VLM semantic scores but measurable on FID/FVD, indicating that stronger identity constraints contribute to both perceptual quality and distributional alignment with the reference set. Neither setting causes metric degradation on any axis, confirming the stability of $\mathcal{L}_{\text{IC}}$. We select $\lambda = 0.5$ as the default for all experiments.

\subsection{Appearance Bank Size $N$}
\label{app:ablation_bank}

The appearance bank $\mathcal{B}$ is constructed by a greedy coverage-set procedure that iteratively adds the training frame covering the most still-uncovered frames within cosine threshold $\tau{=}0.8$. The procedure converges after selecting 123 frames (the full bank). We study how the bank size $N$, i.e., the number of top entries selected from the greedy ordering, affects coverage of the training set. Coverage is measured over all 148{,}040 DINOv2 frame features extracted from the 1{,}900 training videos at a stride of 4.

\begin{table}[h]
\centering
\caption{Effect of appearance bank size $N$ on training-set coverage.}
\label{tab:bank_size}
\begin{tabular}{lcccc}
\toprule
 & $N{=}10$ & $N{=}20$ & $N{=}50$ & $N{=}123$ \\
\midrule
Coverage ($\tau{=}0.8$) & 92.20\% & 96.87\% & 99.51\% & 100.00\% \\
Mean similarity & 0.8532 & 0.8638 & 0.8764 & 0.8861 \\
\bottomrule
\end{tabular}
\end{table}

\begin{figure}[h]
    \centering
    \includegraphics[width=\textwidth]{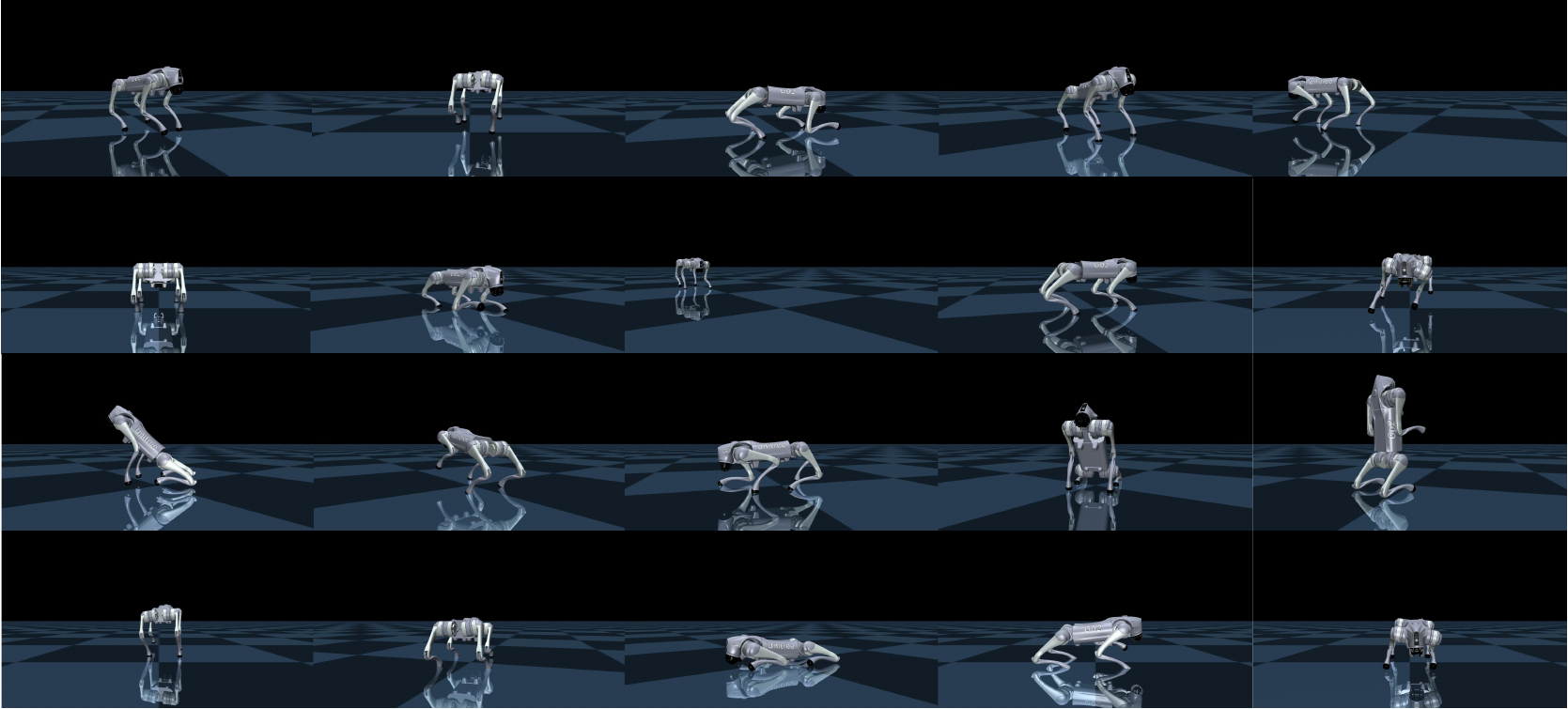}
    \caption{The $N{=}20$ appearance bank frames.}
    \label{fig:bank_entries}
\end{figure}

As shown in Table~\ref{tab:bank_size}, coverage at the operating threshold $\tau{=}0.8$ rises steeply from $N{=}10$ (92.20\%) to $N{=}20$ (96.87\%), with diminishing returns thereafter: $N{=}50$ adds only 2.64 percentage points and the remaining 73 entries contribute a further 0.49 points. Meanwhile, a larger bank increases the per-step cost of the $\max_j$ operation linearly and, more importantly, risks making the hinge loss too easy to satisfy, weakening the identity constraint. We therefore set $N{=}20$ for all experiments, which provides near-complete coverage at $\tau{=}0.8$ while keeping the bank compact enough for the hinge to remain discriminative. The 20 selected bank frames, visualized in Figure~\ref{fig:bank_entries}, span a representative range of viewpoints, body poses, and lighting conditions present in the fine-tuning set.

\subsection{DINOv2 CLS Attention Analysis}
\label{app:dino_attention}

A potential concern with using DINOv2 CLS features for identity consistency is whether the CLS token primarily attends to the robot dog or to the background. If background features dominated the CLS representation, the identity loss would constrain scene consistency rather than subject identity, defeating its purpose.

\begin{figure}[h]
    \centering
    \includegraphics[width=\textwidth]{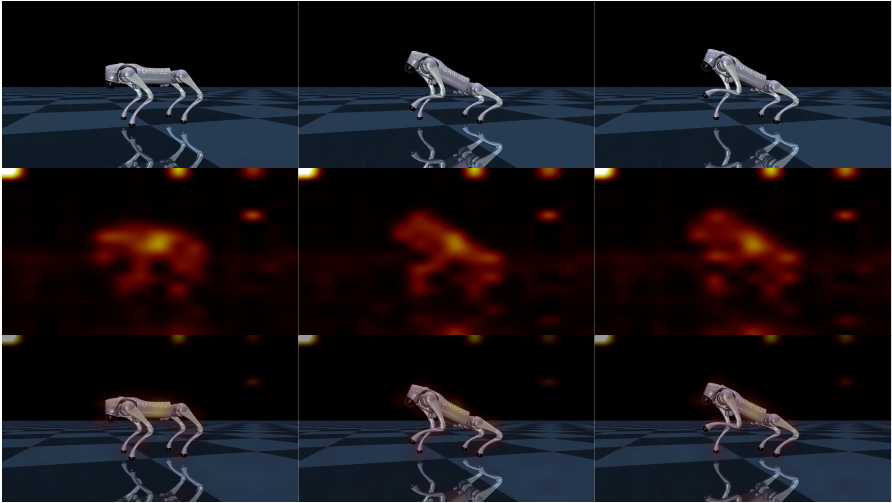}
    \caption{DINOv2 CLS token attention maps on training frames. Top: original images. Middle: attention heatmaps (warmer colors indicate higher attention). Bottom: heatmap overlaid on the original image. The CLS token consistently attends to the robot dog's body and limbs rather than the background, confirming that the CLS feature captures subject identity.}
    \label{fig:dino_attention}
\end{figure}

To verify this, we visualize the self-attention maps of the CLS token in the frozen DINOv2-Base model on representative frames from the training set. Specifically, we average the attention weights from the CLS token to all spatial patches across all heads in the last four transformer layers, reshape the resulting vector into a 2D spatial grid, and upsample it to the original image resolution. The resulting attention heatmaps are overlaid on three example frames with varying backgrounds and poses, as shown in Figure~\ref{fig:dino_attention}.

The attention maps consistently concentrate on the robot dog's torso and limbs across different viewpoints, poses, and background textures, with minimal activation on the floor or surrounding scene. This confirms that DINOv2's CLS representation is inherently subject-centric for our training distribution, and that the identity consistency loss computed from CLS features effectively constrains the robot's appearance rather than background content, without requiring explicit segmentation.

\section{Quad-Imaginarium Dataset Details}
\label{app:dataset}


\paragraph{Per-motion data.} Quad-Imaginarium contains 7{,}488 motion clips totaling 18.5 hours. Each motion is represented as a temporal sequence of per-frame 19-D state vectors in the Unitree Go2 configuration space. The 19 dimensions comprise: 3-D root position in world frame, 4-D root orientation as a unit quaternion, and 12 joint angles in radians, where each of the four legs contributes three joints: hip, thigh, and calf. All trajectories are resampled to 50\,Hz from the original 24\,fps video generation. Clip durations range from 5\,s to 15\,s with a mean of 8.9\,s.

\section{Additional Qualitative Results}
\label{app:qualitative}

\begin{figure}[h]
    \centering
    \includegraphics[width=0.9\textwidth]{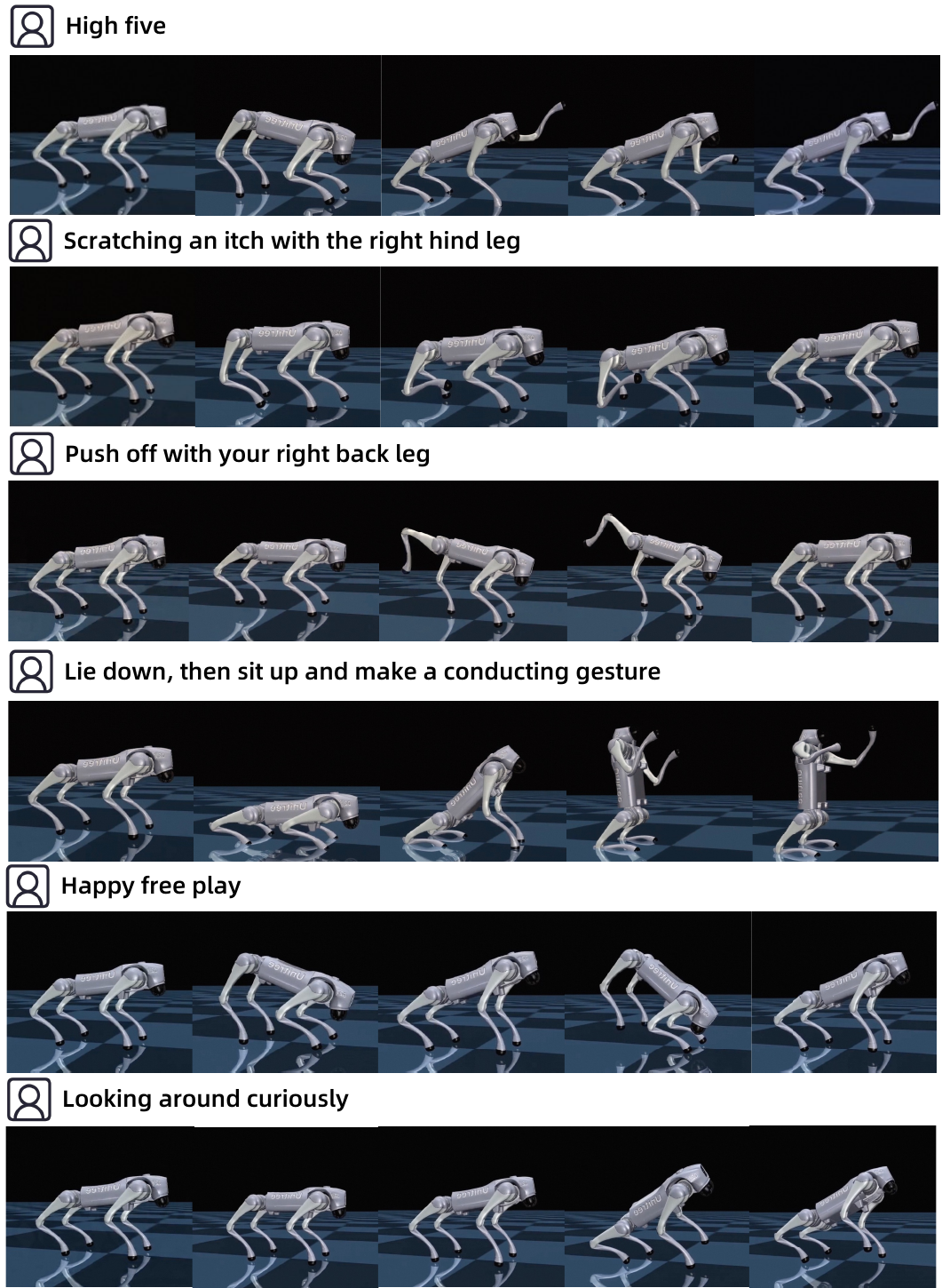}
    \caption{Additional generated video frame sequences from Wan-FT + $\mathcal{L}_{\text{IC}}$, showing six representative motions from Quad-Imaginarium.}
    \label{fig:more_gen}
\end{figure}

\paragraph{Generated video examples.} We present additional frame sequences from our fine-tuned video generation model in Figure~\ref{fig:more_gen}, illustrating the diversity of motions the pipeline can produce. Across all clips, the robot maintains a consistent rigid-body appearance without the melting or distortion artifacts observed in the baseline models.

\begin{figure}[h]
    \centering
    \includegraphics[width=0.9\textwidth]{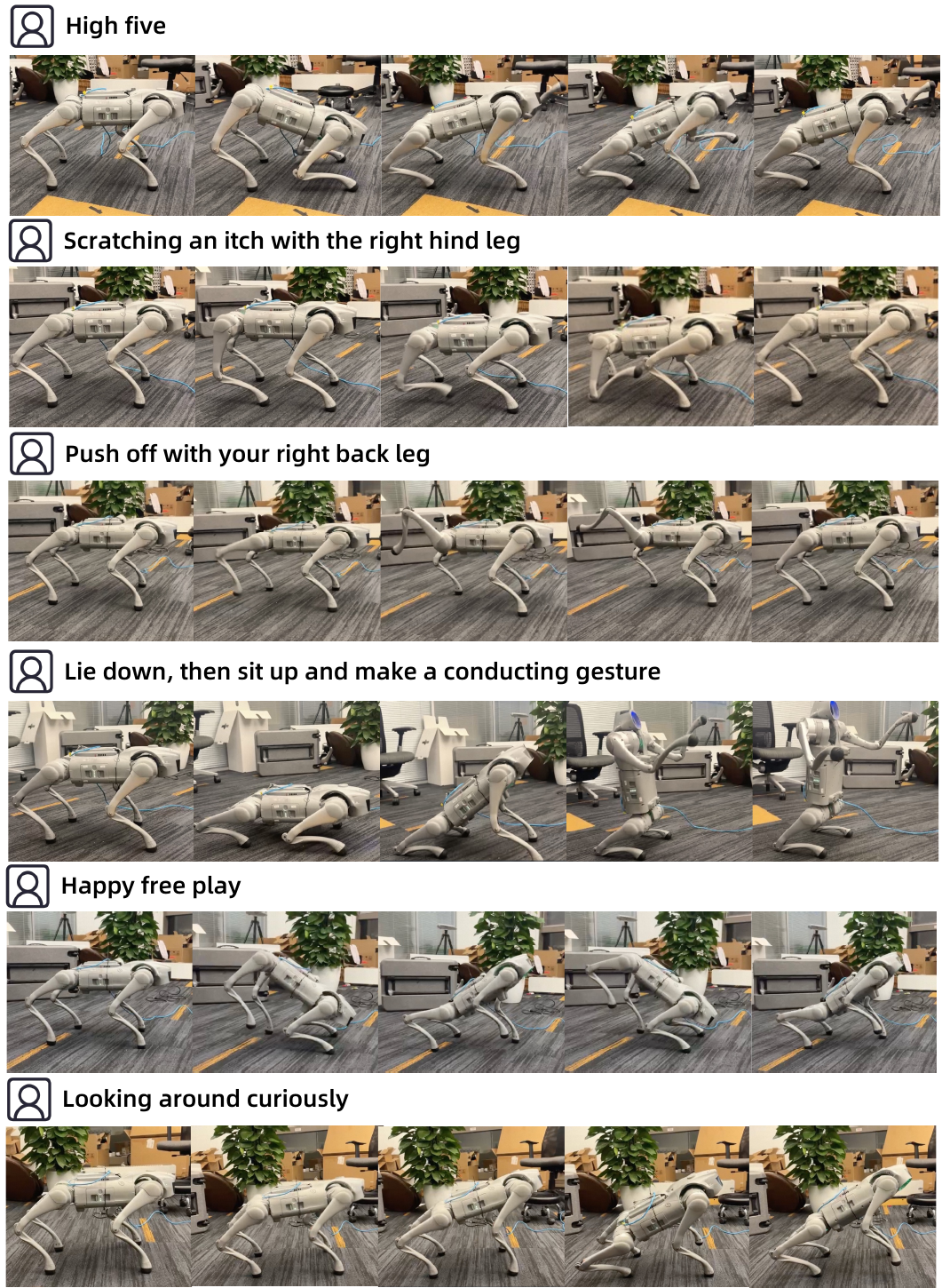}
    \caption{Real-robot executions of the six motions from Figure~\ref{fig:more_gen} on the Unitree Go2.}
    \label{fig:more_real}
\end{figure}

\paragraph{Real robot execution.} The corresponding real-robot deployments of the same six motions on the Unitree Go2 are shown in Figure~\ref{fig:more_real}, all learned entirely from generated video without any animal motion capture data.

\paragraph{Supplementary video.} We provide a supplementary video that compiles real-robot demonstrations of diverse motions from Quad-Imaginarium executed on the Unitree Go2, showcasing the breadth of the learned motion repertoire.

\end{document}